\documentclass{article}

\PassOptionsToPackage{numbers, compress}{natbib}

\usepackage[final]{neurips_2019}
\usepackage[utf8]{inputenc} 
\usepackage[T1]{fontenc}    
\usepackage{hyperref}       
\usepackage{url}            
\usepackage{booktabs}       
\usepackage{amsfonts}       
\usepackage{nicefrac}       
\usepackage{microtype}      
\usepackage{graphicx}       
\usepackage{xcolor}
\usepackage{caption}
\usepackage{subcaption}
\usepackage{float}
\usepackage{wrapfig}
\usepackage[compact]{titlesec}

\title{Objects of violence: synthetic data for practical ML in human rights investigations}

\author{%
  Lachlan Kermode \thanks{Equal contribution.} \\
  Forensic Architecture\\
  \texttt{lk@forensic-architecture.org} \\
  \And
   Jan Freyberg \footnotemark[1] \\
   Element AI \\
   \texttt{jan.freyberg@elementai.com} \\
  \And
   Alican Akturk \\
   Forensic Architecture\ \\
   \texttt{aa@forensic-architecture.org} \\
  \And
   Robert Trafford \\
   Forensic Architecture\\
   \texttt{rt@forensic-architecture.org} \\
   \And
   Denis Kochetkov \\
   Element AI \\
   \texttt{denis.kochetkov@elementai.com} \\
   \And
   Rafael Pardinas \\
   Element AI \\
   \texttt{rafael.pardinas@elementai.com} \\
   \And
   Eyal Weizman \\
   Forensic Architecture\\
   \texttt{ew@forensic-architecture.org} \\
   \And
   Julien Cornebise \\
   Element AI and University College London \\
   \texttt{j.cornebise@ucl.ac.uk} \\
}
\newcommand{\myhref}[2]{#2\footnote{\url{#1}}}

\begin{document}

\maketitle
\vspace{-2em}

\begin{abstract}
\vspace{-1em}
   We introduce a machine learning workflow to search for, identify, and meaningfully triage videos and images of munitions, weapons, and military equipment, even when limited training data exists for the object of interest. This workflow is designed to expedite the work of OSINT ("open source intelligence") researchers in human rights investigations. It consists of three components: automatic rendering and annotating of synthetic datasets that make up for a lack of training data; training image classifiers from combined sets of photographic and synthetic data; and \texttt{mtriage}~\citep{mtriage}, an open source software that orchestrates these classifiers' deployment to triage public domain media, and visualise predictions in a web interface. We show that synthetic data helps to train classifiers more effectively, and that certain approaches yield better results for different architectures. We then demonstrate our workflow in two real-world human rights investigations: the use of the Triple-Chaser tear gas grenade against civilians, and the verification of allegations of military presence in Ukraine in 2014.
\end{abstract}

\section{Introduction}

\subsection{Motivation}

Many recent human rights investigations have relied upon finding a specific object (or object class) among a large number of freely available videos or images. Effective manual search-and-analysis of freely available image material has led to remarkable results in human rights research, including the first charges brought at the International Criminal Court based on public domain evidence \citep{costello2018international}. In the Safariland case, although the export of military equipment from the US is commonly a matter of public record, the sale and export of tear gas is not. As a result, it is only when images of tear gas grenades appear online that monitoring organizations and the public can know where they have been sold, and who is using them. The discovery, identification, and archiving of munitions that appear in online images is therefore essential for the pursuit of corporate and government accountability in the global arms trade.

\setlength{\intextsep}{0pt}%
\begin{wrapfigure}{r}{0.5\textwidth}

    \centering
    \begin{subfigure}{.24\textwidth}
        \centering
        \includegraphics[width=\linewidth]{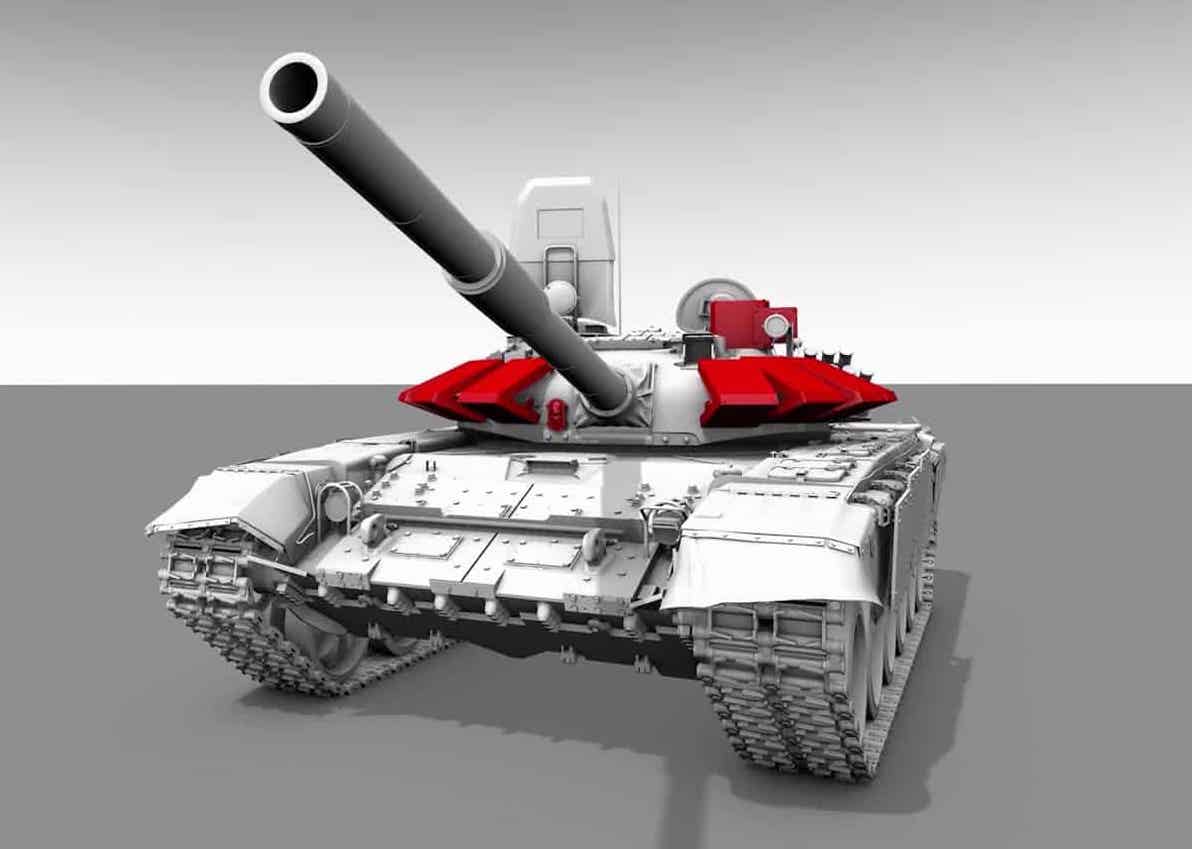}
    \end{subfigure}
    \begin{subfigure}{.24\textwidth}
        \centering
        \includegraphics[width=\linewidth,trim={0 1.8cm 0 0},clip]{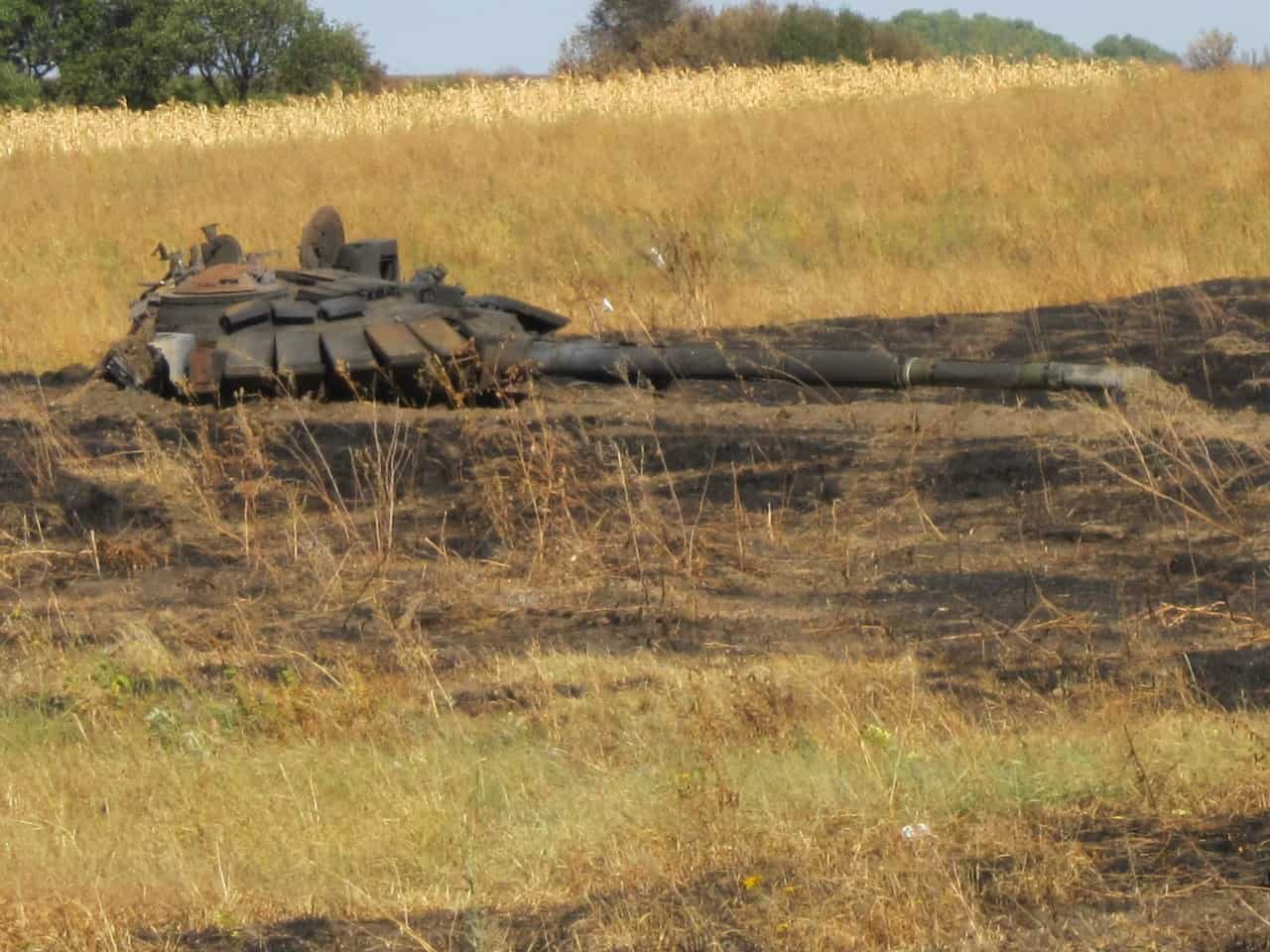}
    \end{subfigure}
    \begin{subfigure}{.24\textwidth}
        \centering
        \includegraphics[width=\linewidth]{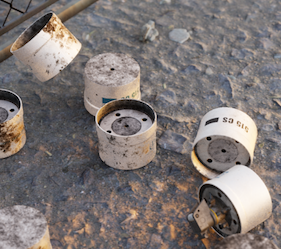}
    \end{subfigure}
    \begin{subfigure}{.24\textwidth}
        \centering
        \includegraphics[width=\linewidth,trim={0 2.1cm 0 2.1cm},clip]{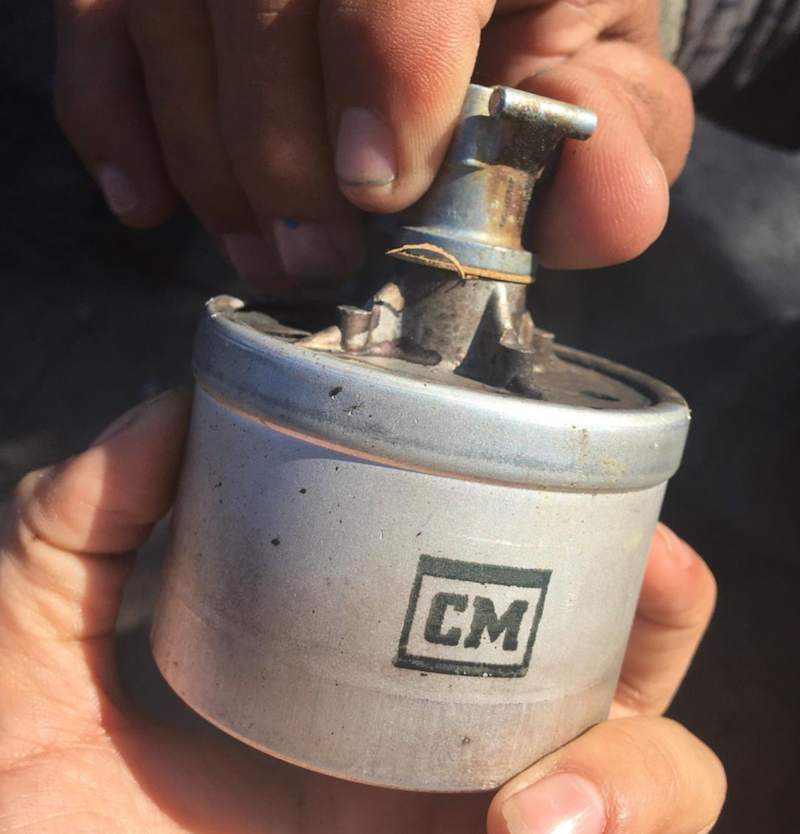}
    \end{subfigure}
    \caption{Our 3D model of a Russian tank (top left) and real turret of a destroyed tank (top right); Our 3D model of the Triple-Chaser grenade (bottom left) and photo of real parts (bottom right).\label{fig:3d}}
    \vspace{-1em}
\end{wrapfigure}

Yet for all its potential impact, searching for images of tear gas online manually is laborious and time-consuming. Automating discovery, identification, or archiving, could be hugely beneficial to human rights monitors. The deployment of computer vision classifiers holds substantial promise for the field of human rights investigation and monitoring at large.

While deep learning is a state-of-the-art technique for computer vision \citep{lecun2015deep}, it is extremely data-intensive. Effective classifiers usually require thousands of annotated training images, even when using pretrained models. Not only is this annotation of real-world images costly and laborious; building such a set relies on there being enough images available (with appropriate licensing) in the first place. When investigating the use of Triple-Chaser, a tear gas canister manufacted by a group called Safariland, we found that public images of the Triple-Chaser are relatively rare.  Synthetic datasets of different flavours have been used in various domains to create supplemental training data \citep{miglioAIUnrealEngine, tremblayTrainingDeepNetworks2018, rosSYNTHIADatasetLarge2016, tremblayFallingThingsSynthetic2018}. The authors know of one other initiative to use synthetic data for the purpose of documenting human rights violations, VFRAME \citep{vframeio}.

\subsection{Contribution}
To supplement a lack of training data in various cases, we developed an approach to generate annotated synthetic training data sets. Our synthetic data is generated using both Unity and Unreal Engine. We use a combination of purely artificial 'low-fidelity' images, and 'high-fidelity' ones that recreate the visual contexts in which the object of interest has been previously documented. We illustrate the potential of synthetic data by looking in detail at the Triple-Chaser case, providing a comparative study of the advantage of synthetic data with various model architectures and training methods. 

To scale inference across a range of public domain media, a requirement for a classifier's meaningful use in human rights research, we developed a workflow to apply classifiers on the frames of videos scraped from Youtube. We also developed a web interface to allow human rights researchers to interactively visualise predictions, and thus streamline results as a part of OSINT research. We release the tool \texttt{mtriage} as open-source \citep{mtriage}.

Finally, we report on the impact this work has had on our human rights investigations.

\section{Generating synthetic data}
By carefully studying images collected in the field, we used architectural modelling and game engine software software to generate annotated, high-fidelity synthetic datasets.

\subsection{Domain randomisation: parametric modification of 3D models and render perspective}

Referencing a photogrammetric reconstruction of the Triple Chaser (see \ref{sec:photogrammetry}) alongside other photographs, we used Adobe's Substance Designer \citep{SubstanceDesigner} to create parametric photo-realistic textures for the grenade. In \ref{sec:unreal-appendix}, we outline our reasons for choosing Unreal. In the Substance texture, the branding color and content for the grenade, which varies around the world, can be modified at render time to produce a range of variations. We can similarly modify weathering and wear effects on the canister, such as dust, dirt, grime, scratches, and bends.

We then insert the 3D model into a scene that randomly varies the render camera's position, as well as modifying settings such as exposure, focal length, and depth of field, between frames. After rendering each frame, we also apply post-effects such as LUTs for color correction, variations in the film grain, and different vignettes. See Appendix~\ref{sec:texture-variations} for demonstrative images. In a parallel stream of work, we also developed a Unity 3D pipeline that renders images at lower fidelity (see \ref{fig:boxes-and-masks} but contains distractor objects that further randomise the viewing conditions the Triple Chaser grenade appears in.

\subsection{Producing annotated synthetic datasets}

Using Unreal, we produced thousands of synthetic images depicting the Triple-Chaser in various conditions. Following  Moore's approach \citep{nespressoBlog}, the grenade was depicted both against "decontextualized" patterned backgrounds as well as simulated real-world environments. In addition to each rendered image, we also rendered an additional `mask' for each image that contains pixel-level annotations. By comparing this mask with the original image, we then generated bounding boxes and bitmap masks for each rendered image (see \citep{ue4supervisely}). See Appendix~\ref{sec:boxes-and-masks} for demonstrative images.

\section{Using synthetic data}

\subsection{Numerical evaluation on Triple Chaser}

We evaluate the utility of synthesised images containing the Triple Chaser on a small data set of real, labelled images containing the grenade. We trained three separate models: a classification network, a semantic segmentation network, and an instance segmentation network. For details of the training data and the model architectures, see \ref{sec:models-and-training}.

Training with synthetic data increased performance of the model. In particular, training with synthetic data allowed models to train significantly faster. We observed that training with domain adversarial methods provided significant performance increases when using classification models, while that increase was smaller when using a semantic segmentation model, and nonexistent with a mask model.

We believe that this could be alleviated with a more sophisticated domain adaptation approach. The domain adaptation achieved with our architecture only affected roughly half the parameters in the U-Net architecture, and an even smaller proportion in MaskRCNN, which may explain why it didn't produce a larger up-lift in those models. On the other hand, mask predictions encourage models to process information locally and retain visual information necessary to identify the Triple Chaser grenades, which may mean that domain adaptation is less necessary. This might be because the distance between synthetic and real distributions of pixels on the canister is much lower than the distance between distribution of pixels in the background.

\begin{figure}[H] 
    \centering
    \includegraphics[width=1\linewidth]{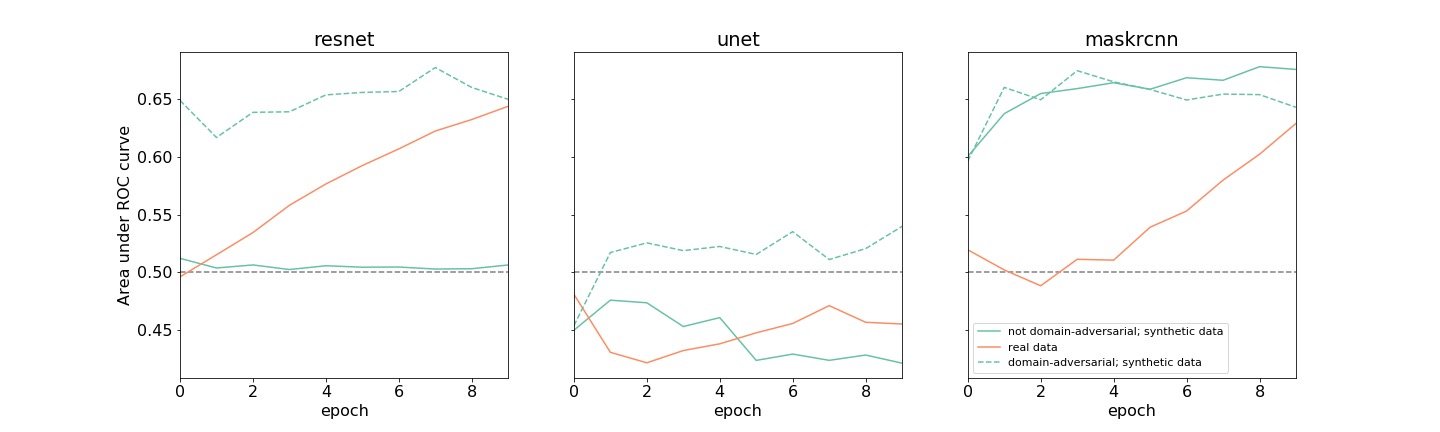}
    \caption{Area under ROC curve of models trained for 10 epochs. For details of the training procedures and data used, see \ref{sec:models-and-training}. Synthetic data provides faster learning, and effects of domain difference can be alleviated with domain adaptation.}
    \label{fig:eval}
    \vspace{-1em}
\end{figure}

It is important to note the different standards by which such applications are useful. Because there are direct legal implications related to the discovery of material, neural network output must be verified by human experts. Our image classifier's task is therefore to narrow down the number of images for the human expert. Evaluation of this work purely with classifier accuracy does not assess the usefulness of the workflow in practice. Thresholds for network predictions are set dynamically: after ranking based on network scores, humans simply start at the top and work their way down until they realise the predictions are no longer accurate. For this reason, we believe that area under either ROC or P/R curves are informative; but that user studies of OSINT researchers interacting with the model output surfaced via mtriage would produce better evaluation metrics.

\subsection{Analysing media at scale with Mtriage}

The bulk of media used during human rights investigations derive from major platforms such as Youtube, Twitter, and Facebook. These platforms often have robust APIs, but they differ in various ways: in the authentication they require, the search parameters they make available, and the media formats they support. We introduce Mtriage, a modular open source software, to produce a standardised media format across various platforms, and to effectively orchestrate analysis at scale across this media. Mtriage contains two types of components: \emph{selectors} to search for and download media, and \emph{analysers} to derive data from selected media. See Appendix~\ref{sec:mtriage} for more detail regarding Mtriage's architecture and use-case.

In this work, we use Mtriage to run inference at scale, utilising a NVIDIA-RTX 2080Ti GPU and running in parallel on 16 CPUs, using classifiers trained with synthetic data to make predictions on videos scraped from Youtube. The workflow splits each video into frames sampled at a rate of one frame per second, and then runs inference with the classifier for each sampled frame, predicting the likelihood of a grenade in each image. The predictions are then visualized in a web interface (also shown in Appendix~\ref{sec:mtriage}).

\subsection{Real-world use in human rights investigations}

In the Triple-Chaser investigation, a fully trained human rights researcher spent one full-time week to crawl online sources and surface merely 173 images of the grenades. Our pipeline Mtriage scanned the same sources in a couple of hours.

In the investigation "The Battle of Ilovaisk" \citep{TheBattleofIlovaisk}, we experimentally ran Mtriage to collect images of tanks using a ResNet50 classifier trained on ImageNet \citep{deng2009imagenet}. Previously, an experienced open source researcher who is an expert in Ukraine/Russia relations had worked full-time for three weeks to collect some 37 videos from Youtube that he deemed relevant.

We ran the Mtriage workflow using three or four search terms related to the conflict (a far narrower net than that which the researcher had cast while searching manually), which yielded approximately 600 videos. Download, frame extraction, and inference on these took approximately 4 hours, running on a single computer, leveraging GPU capabilities via CUDA where appropriate. \thanks{Workflows in this paper were run in serial, as Mtriage operations had not been parallelized at the time.} After sorting these videos by the relative volume of frames with a positive prediction, in less than 20 minutes the same researcher found two previously undiscovered videos that were instrumental in the research more generally.

While these two single cases cannot substitute a varied and more extensive study, Mtriage's use in these investigations does indicate our pipeline's powerful potential in supporting specialised researchers to expedite media discovery in human rights research.

\section{Future work}
\label{sec:future-work}
This research shows that synthetic data is a valuable addition to a researcher's toolkit when training classifiers. It will be exciting to further qualify the potential of synthetic data by experimenting with varied and additional domain randomization techniques in Unity, Unreal, and other graphics engines. 

The early enthusiasm for this research from our partners in the fields of OSINT research and human rights investigation and monitoring has been very encouraging. Through workshops and ongoing consultation with organisations such as Amnesty International, its Citizen Evidence Lab \citep{CitizenEvidenceLab} and Digital Verification Corps, Bellingcat, Human Rights Watch, the UN's Office for the High Commissioner of Human Rights, and the Syrian Archive, we are in the process of co-conceiving new use cases through which both this research can be furthered, and the workflow further tested and improved. We are actively developing a project to create an AI model zoo for human rights research, so that techniques in computer vision can be further democratised across many flavours of human rights research using Mtriage and other tools.

\makeatletter
\if@submission
\else
\subsubsection*{Acknowledgments}
We thank Emily Jacir, Sarit Michaeli, Oren Ziv, Zuhal Altan, Micol (@\_EN\_Mexico on Twitter), Phoebe Cottam, and Giedre Steikunate for contributing images of the Triple Chaser. We would also like to acknowledge Praxis Films for working closely with Forensic Architecture to produce the film that was shown in the 2019 Whitney Biennial documenting this research, as well as David Byrne, Anna Feigenbaum, Neil Cormey and Adam Harvey. We thank Alex Kuefler from Element AI for some of his computer vision code. Frank Longford conducted a six-week study on the efficacy of synthetic data as a research fellow at Forensic Architecture in 2018, thanks to the generous support of Faculty data science; that work was a prelude to the much of the research presented in this paper.
\fi 
\makeatother

\raggedbottom
\bibliography{bibliography}

\pagebreak
\appendix

\section{Appendix}

\subsection{Photogrammetry}
\label{sec:photogrammetry}
\begin{figure}[H] 
    \centering
    \begin{subfigure}{.45\textwidth}
        \centering
        \includegraphics[width=\linewidth]{images/image15.png}
    \end{subfigure}
    \begin{subfigure}{.45\textwidth}
        \centering
        \includegraphics[width=\linewidth]{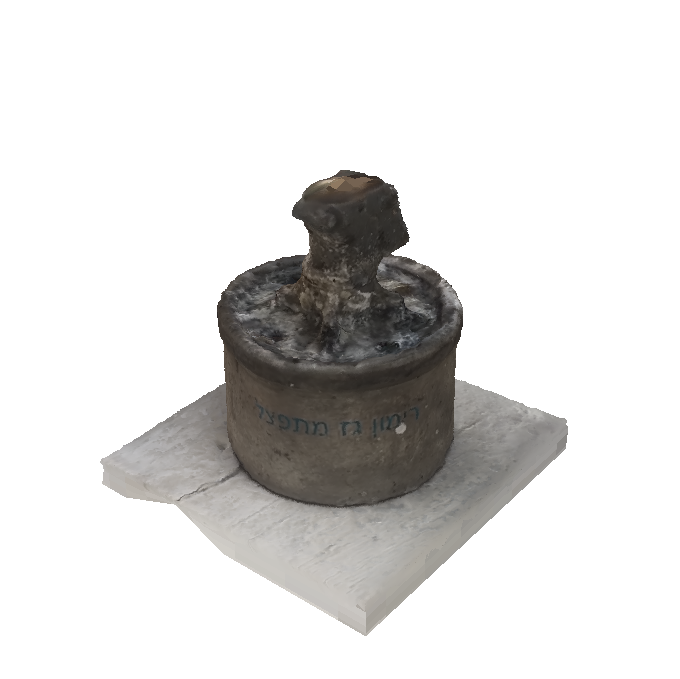}
    \end{subfigure}
    \caption{Photogrammetry is a Structure-from-Motion algorithm that allows the reconstruction of 3D assets given multiple photographic perspectives of an object as input.}
    \label{fig:photogrammetry}
\end{figure}

\subsection{Unreal Engine}
\label{sec:unreal-appendix}

We could have worked in a number of 3D frameworks, but we chose Unreal for a number of reasons:

\begin{enumerate}
\def\labelenumi{\arabic{enumi}.}
\item
  \textbf{Purpose-built for dynamic perspective}. In
  \myhref{https://engineering.forensic-architecture.org/experiments-in-synthetic-data}{previous
  research}, we built an experimental framework in
  \myhref{https://www.maxon.net/en/products/cinema-4d/overview/}{Maxon's
  Cinema4D} for this same purpose. Cinema4D is a general purpose
  modelling and animating suite, similar in principle to
  \myhref{https://www.blender.org/}{Blender} or
  \myhref{https://www.autodesk.com/products/maya/overview}{Maya}. While
  there are certain advantages using it, when using it we found that we
  were required to script many camera, texture, and model variations
  from scratch. Indeed, many of the functions we require for a synthetic
  data generator-- such as domain randomization, parametric textures,
  and variable lighting-- are more common in game development than in
  animation. Unreal offers a lot of relevant functionality `out of the
  box', and has a rich online support community.
\item
  \textbf{Powerful visual scripting}. Unreal's node-based visual
  scripting language
  \myhref{https://docs.unrealengine.com/en-us/Engine/Blueprints}{blueprints}
  is more accessible than code for the architects, game designers, and
  filmmakers with whom we work. We used Unreal blueprints to programmatically modify camera position, camera settings (exposure, focal length and depth of field), as well as image post-effects (LUT's for color correction variations, film grain, and vignette).
\item
  \textbf{Full source code access}. Unreal's source code is available in
  its entirety (although you need an Epic Games account in order to view
  Github repository.) It's not exactly `open source', as the license
  under which it is released does not allow for redistribution or
  adaptation, but source code access allows us to understand performance
  bottlenecks, and to see the full scope of what is possible in the
  software.
\item
  \textbf{Real-time raytracing}. Unreal's capability for real-time
  rendering is the cherry on the top of Unreal's suitability for
  synthetic data generation. The real-time quality of rendering means
  that, in contrast to a workflow using Cinema4D, there are much
  shorter turnarounds in the ongoing conversation between generating and
  training phases.
\end{enumerate}

\subsection{Texture variations}
\label{sec:texture-variations}
\begin{figure}[H] 
    \centering
    \begin{subfigure}{.2\textwidth}
        \centering
        \includegraphics[width=\linewidth]{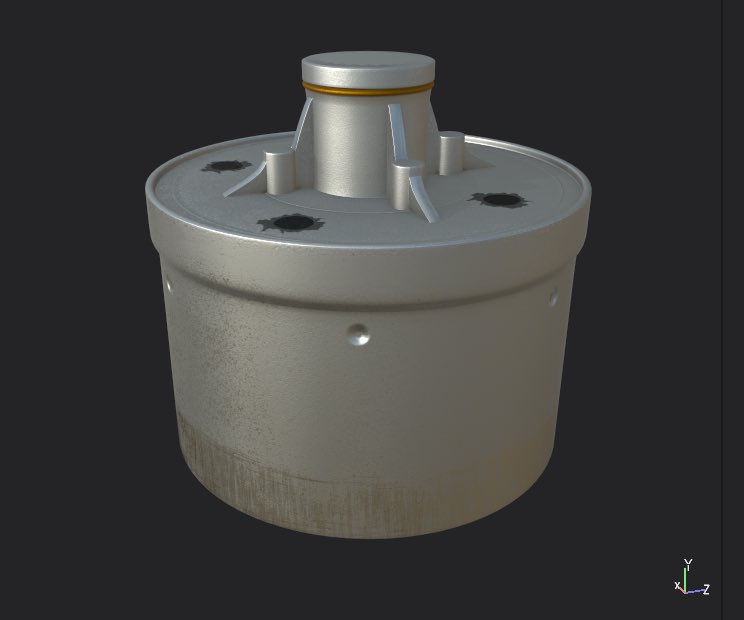}
    \end{subfigure}
    \begin{subfigure}{.2\textwidth}
        \centering
        \includegraphics[width=\linewidth]{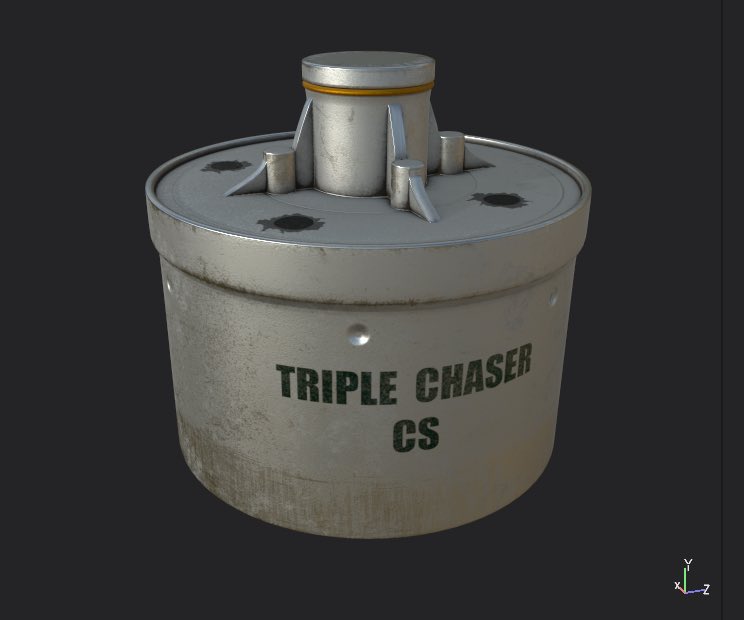}
    \end{subfigure}
    \begin{subfigure}{.2\textwidth}
        \centering
        \includegraphics[width=\linewidth]{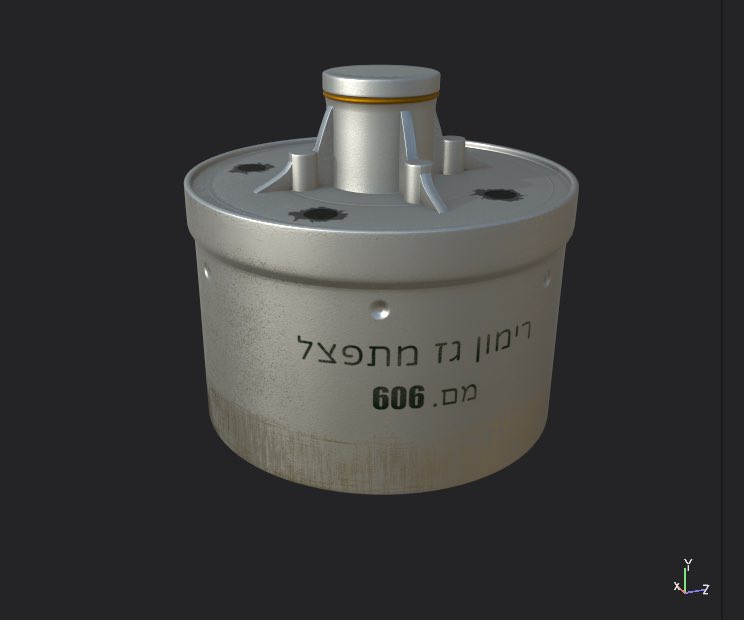}
    \end{subfigure}
    \begin{subfigure}{.2\textwidth}
        \centering
        \includegraphics[width=\linewidth]{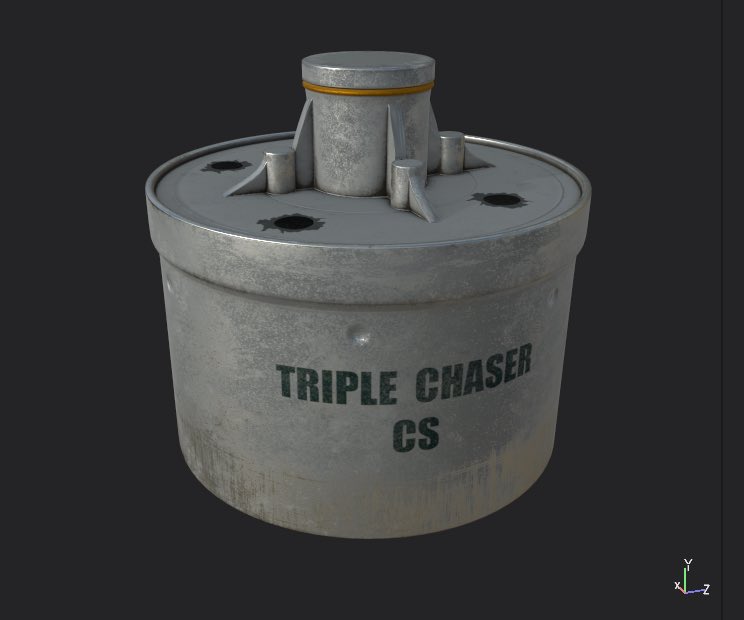}
    \end{subfigure}
    \begin{subfigure}{.2\textwidth}
        \centering
        \includegraphics[width=\linewidth]{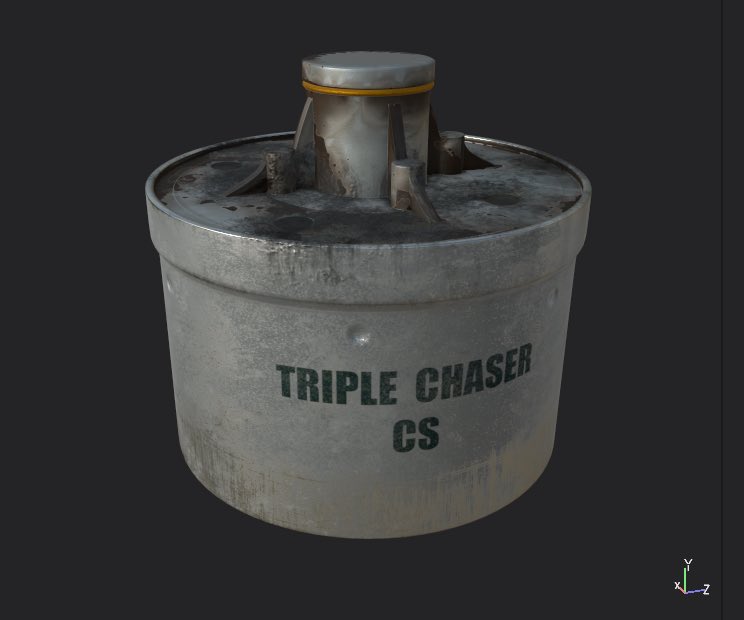}
    \end{subfigure}
    \begin{subfigure}{.2\textwidth}
        \centering
        \includegraphics[width=\linewidth]{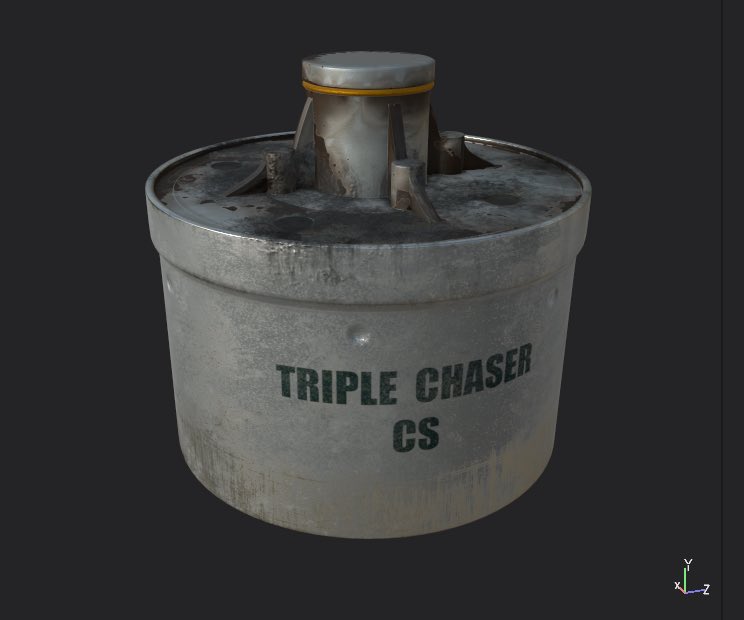}
    \end{subfigure}
    \begin{subfigure}{.2\textwidth}
        \centering
        \includegraphics[width=\linewidth]{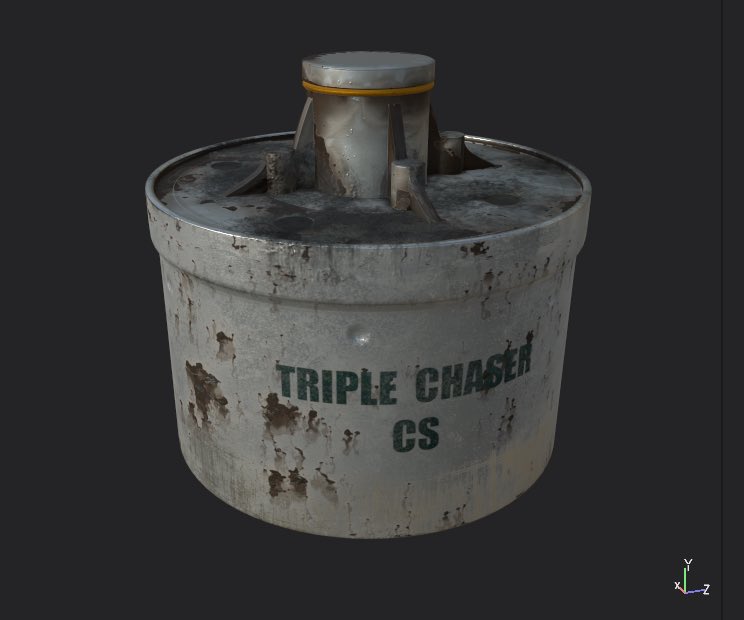}
    \end{subfigure}
    \begin{subfigure}{.2\textwidth}
        \centering
        \includegraphics[width=\linewidth]{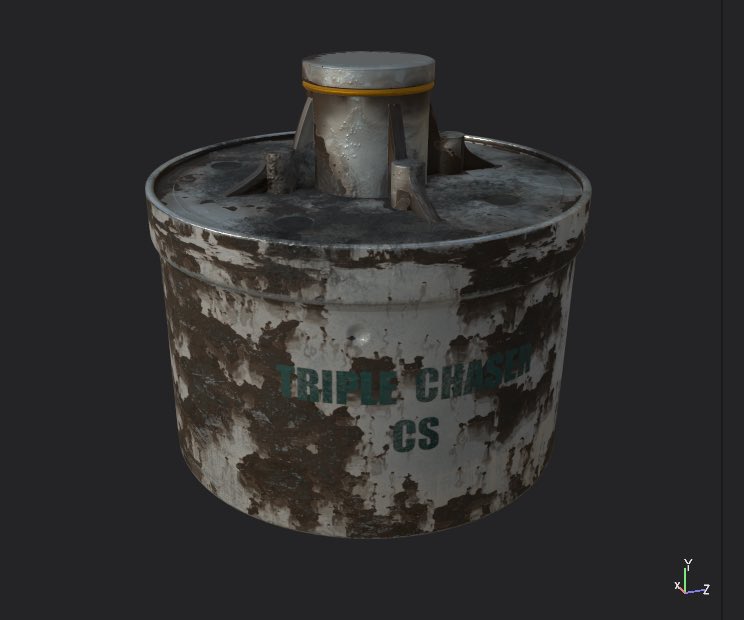}
    \end{subfigure}
    \caption{Various amounts of grit and dirt on a rendered model of the Triple Chaser (increasing in intensity from left top to right bottom). The model of the grenade is represented with no label (first image from left top) and with a varied label (third image from left top).}
    \label{fig:texture-variations}
\end{figure}

\subsection{Rendering less photorealistic images in Unity}
\label{sec:unity}

In addition to data rendered in UE4, we also rendered data with the Unity 3D engine. We used the same 3D models and the same parametric variations of the Triple Chaser grenade in Unity. However, we added domain randomisation features to our rendering engine. In particular, we sampled background images from the Flicker 8K dataset \cite{flickr8k}, and added "flying distractors" to the scene. These geometric objects (cubes, spheres, cylinders) added additional lighting complexity to the dataset and allowed for canisters to be partially obscured.

\subsection{Mask annotations}
\label{sec:boxes-and-masks}
\begin{figure}[H] 
    \centering
    \begin{subfigure}{.45\textwidth}
        \centering
        \includegraphics[width=\linewidth]{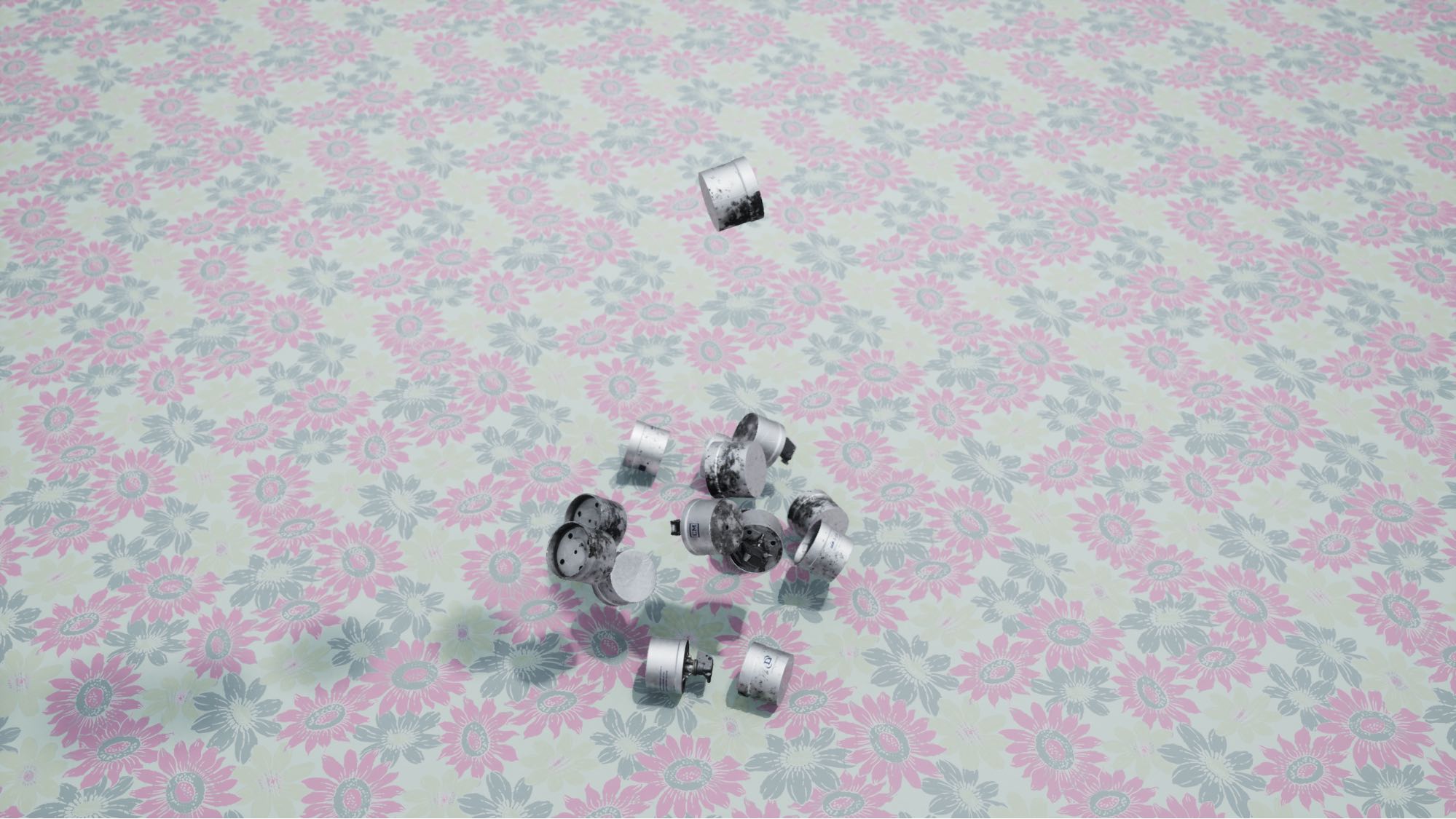}
    \end{subfigure}
    \begin{subfigure}{.45\textwidth}
        \centering
        \includegraphics[width=\linewidth]{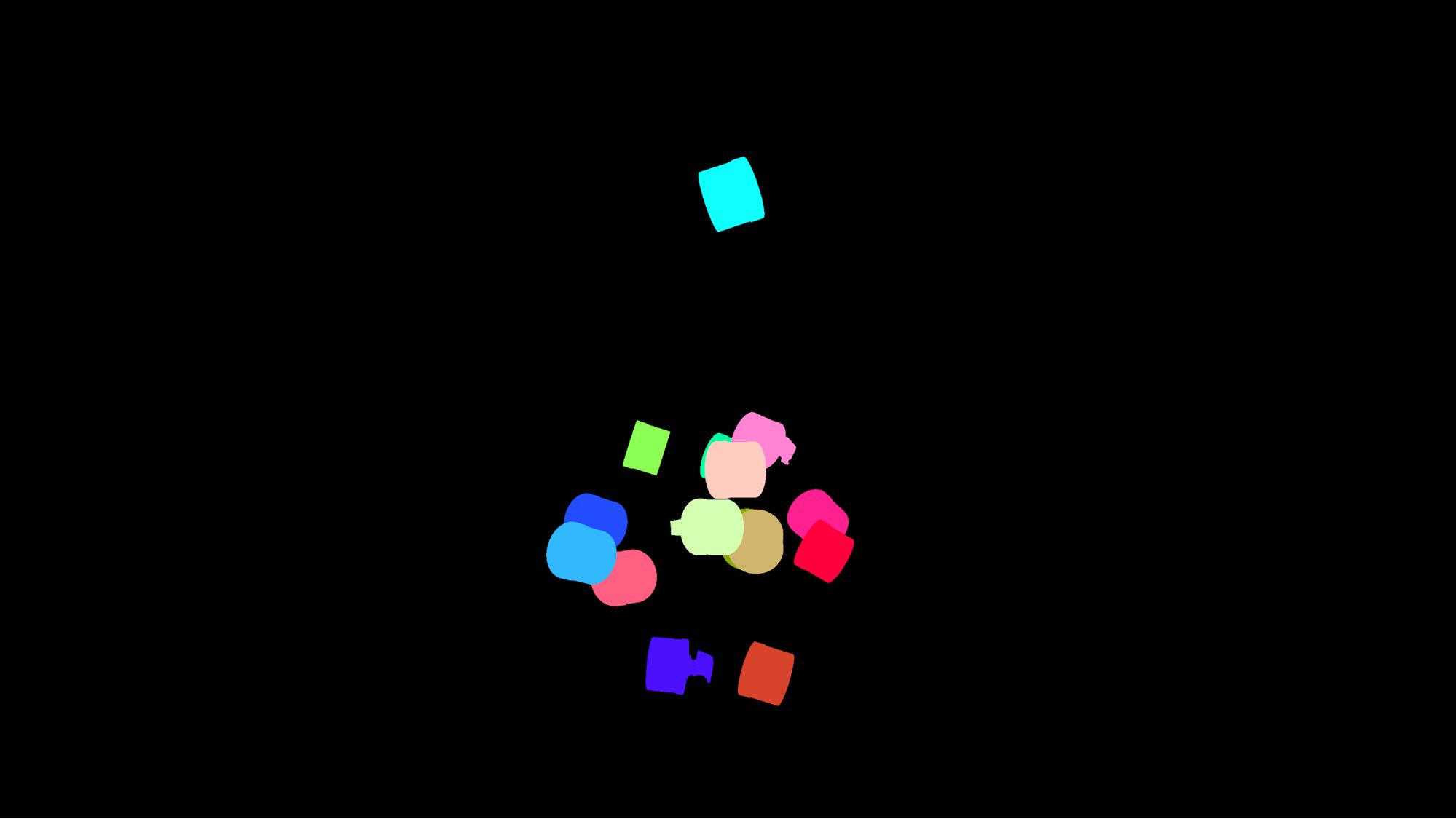}
    \end{subfigure}
    \begin{subfigure}{.45\textwidth}
        \centering
        \includegraphics[width=\linewidth]{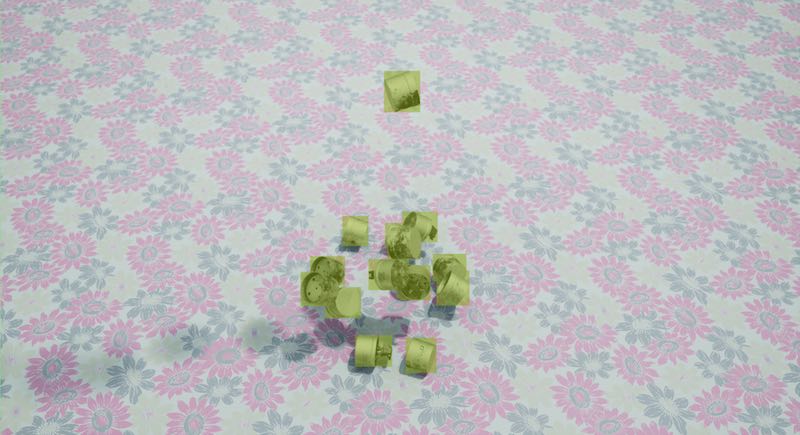}
    \end{subfigure}
    \begin{subfigure}{.25\textwidth}
        \centering
        \includegraphics[width=\linewidth]{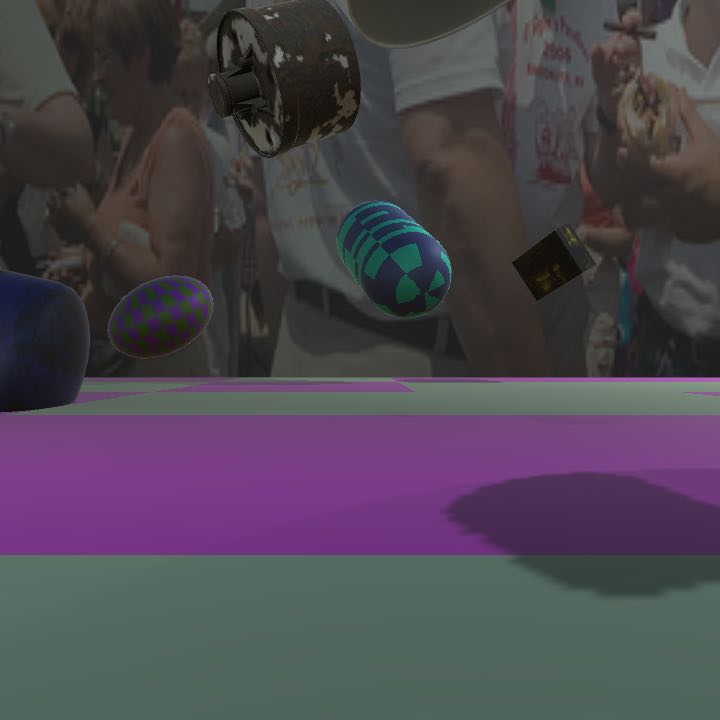}
    \end{subfigure}
    \caption{A rendered synthetic image (left top), the corresponding pixel-level instance mask annotation (right top), and the training image with bounding boxes overlaid (bottom). Images were also rendered with Unity at less high-resolution, and with less photorealistic textures and render settings, but with flying distractors and randomised photographic backgrounds.}
    \label{fig:boxes-and-masks}
\end{figure}

\subsection{Mtriage}
\label{sec:mtriage}
\begin{figure}[H] 
    \centering
    \begin{subfigure}{.9\textwidth}
        \centering
        \includegraphics[width=\linewidth]{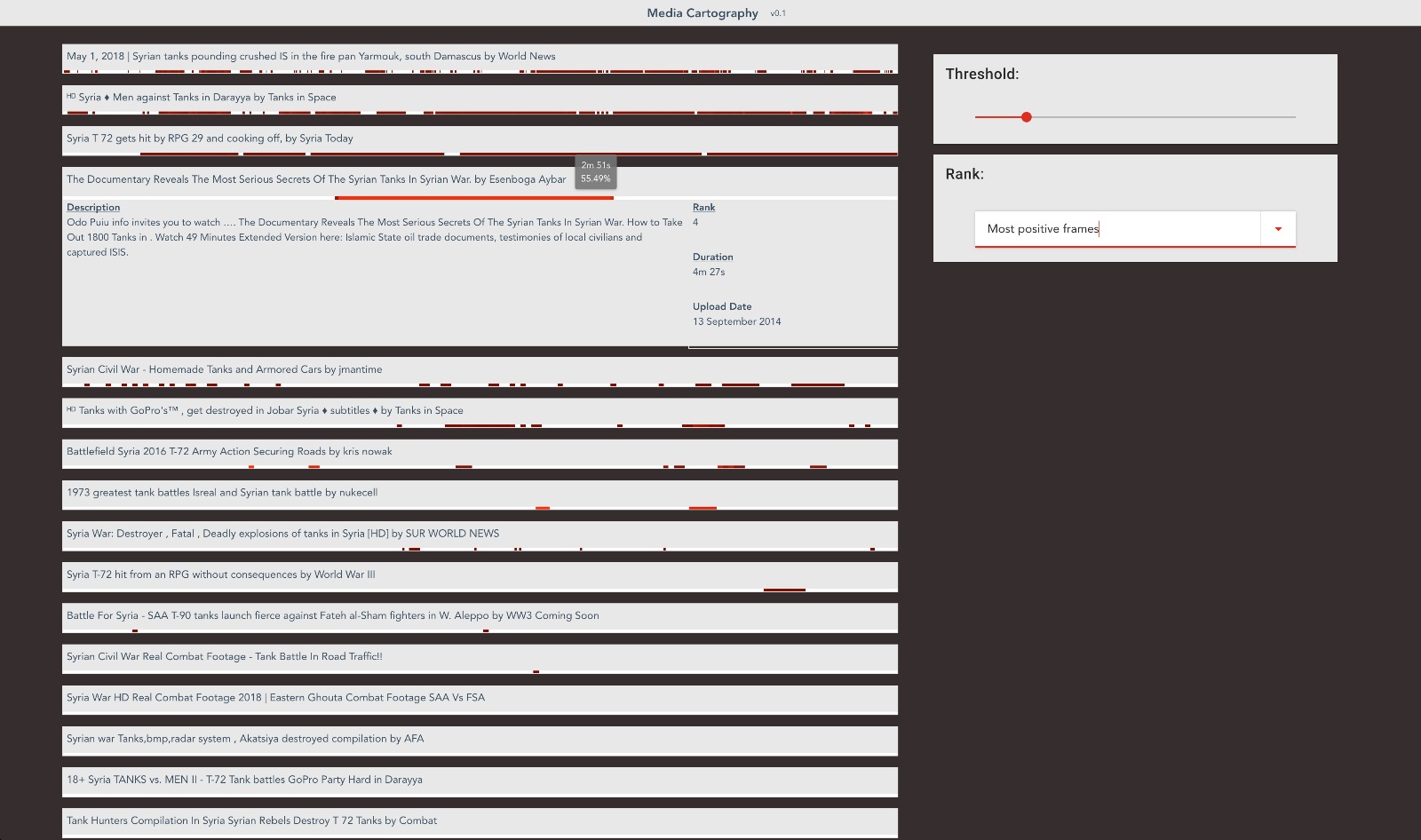}
    \end{subfigure}
    \caption{The web interface presented by mtriage-viewer to the visualise results of object detection inference on the frames of Youtube videos, processed in mtriage. Each grey cell represents the timeline of a video. Red frames in the videos timeline represent positive predictions, with brighter red indicating a higher confidence, and darker red the opposite. When the user hovers over a section in the timeline, the exact confidence is displayed in a tooltip. Controls to apply a minimum threshold at which to display a prediction (i.e. 25\%) and to rank videos by different metrics are available on the right.}
    \label{fig:mtriage}
\end{figure}

As we increasingly found ourselves writing repetitive logic to scrape, download, preprocess, and process media, we developed Mtriage. At a high level, Mtriage is a command-line interface to orchestrate computations that read and write to the local filesystem. Computations are implemented in Python, and are inserted into Mtriage's lifecycle through two related abstractions: \textbf{selectors}, which sample media from various platforms and comport them as Mtriage 'elements' (media with well-defined types on the filesystem); and \textbf{analysers}, which process pre-selected elements, and create corresponding elements.

The base requirement for an \verb|Analyser| is an implementation for the \verb|analyse_element| function, which expresses the computation performed to read an element and produce a new one. The analysis performed in an analyser is conceived as a one-to-one function that reads an element from the filesystem, and eventually writes another. Each analyser has self-contained dependencies, as each is run in a separate Docker container, with Mtriage mounting the appropriate parts of the filesystem as volumes during runtime. Analyser code is arbitrarily complex and, though it must be invoked via Python, can execute logic in any language that can be virtualized in Docker (all languages). This architecture allows developers to architect complex orchestration workflows, treating each analyser as a self-contained microservice. When running analysis, Mtriage scales \verb|analyse_element| to operate across N elements, running operations in parallel according to the number of CPUs available on the host computer. Other class functions such as \verb|pre_analyse| and \verb|post_analyse| can be optionally implemented if setup or teardown (logic that does not need to be run for every element, but only once) is necessary.

The \verb|Selector| class is quite similar to \verb|Analyser|, but differs in that it does not take elements as input, but instead generates elements by retrieving media based on the configuration variables passed. The Youtube selector, for example, requires a search term and start/finish dates, and generates elements that contain the videos uploaded between those dates that are returned from a Youtube search. Mtriage assumes that the \verb|retrieve_element| function contains some asynchronous logic, and thus both parallelizes operations across muliple CPUs, and cooperative multitasking on each CPU.

The selector or analyser to be used, as well as the configuration specific to that component, is written in a single YAML configuration file that is passed to Mtriage via its path from the command line. Logs are both printed to the console and kept on the local filesystem for reference once a component's process is complete. If a process is interrupted for any reason, and Mtriage run again with the same configuration file, the process will continue from where it left off, only running logic on elements that have not yet been processed. Selectors and analysers can be chained together by a unique \verb|meta| analyser that works as a higher-order function to chain the logic of multiple components into one.

Once a process has been run to completion using Mtriage, the result can be effectively visualized using \textbf{mtriage-viewer}, an auxiliary tool we developed and which is available at \myhref{https://github.com/forensic-architecture/mtriage-viewer}{https://github.com/forensic-architecture/mtriage-viewer}. 

For more detail and instructions on how to use mtriage in your own research, see \myhref{https://github.com/forensic-architecture/mtriage}{https://github.com/forensic-architecture/mtriage}. Mtriage is available under the DoNoHarm license on Github. We welcome community pull requests and new contributors.

\pagebreak
\subsection{Details of training architectures of varying complexity on synthetic images.}
\label{sec:models-and-training}

During our model training process, our synthetic training data consisted of images rendered using different parameters and tools. In total, we used 700 high-fidelity images rendered with UE4, and 20,000 images rendered with low-fidelity using Unity. For parametric variations in rendering, see \ref{sec:texture-variations}.

We combined our synthetic data with images not relevant to the task from the Pascal VOC dataset. Each model was trained to predict the Triple Chaser class, as well as the 20 "nuisance" classes in Pascal VOC which were ignored during evaluation. Using these real images enabled us to train a model using domain adversarial training.

During training, images were sampled equally from the low-fidelity, high-fidelity, and Pascal VOC datasets. In practice, this meant that we significantly under-sampled from the low-fidelity images.

We trained three different architectures on this task: a resnet 50 classifier \cite{DBLP:journals/corr/HeZRS15}, a u-net \cite{DBLP:journals/corr/RonnebergerFB15} with Resnet 50 as the encoder, and a mask-rcnn model \cite{maskrcnn_he_2017} with resnet 50 as the backbone in the feature pyramid network. For full implementation details, see the citations for each architecture. When calculating performance, to obtain image-level class predictions, we used the softmax probabilities from the classifier, the average of the top-5\% of pixelwise softmax probabilities from the semantic segmentation network, and the maximum class probability of the MaskRCNN after non-max suppression.

In all cases, we trained networks using the Adam optimizer with a learning rate of 1e-4 and L2 regularisation of 1e-6. Domain adversarial training was achieved by extracting intermediate layer output (layer four of Resnet 50) from the model and processing it using a simple fully connected classifier trained to discriminate between 3 domains: real-world photography, low-fidelity synthetic images (i.e. generated with Unity), and high-fidelity synthetic images (i.e. UE4). This discriminator uses linear layers with ReLU activation. Since Resnet50 was used in all our architectures, it processed the output of layer 4 (2048 feature maps), with 3 fully connected layers (with 512, and 256 units in hidden layers and 3 output units, respectively). We optimized cross-entropy of class predictions in these fully connected layers, but reversed the gradients backpropagated into the main, task-specific network. Moving weights in the opposite direction to the gradient that optimises domain classification leads to domain adaptation: parameters become optimised to produce intermediate features that are indistinguishable between domains \cite{ganin2016domain}. To stabilise training, we multiplied the domain adversarial loss with a weighting alpha, which we varied between 0.001, 0.05, and 0.1.

This implementation of domain adversarial training was originally designed for classification models. The reversed gradients only flow through parts of the feature representation in the segmentation networks and therefore only affect a small proportion of the parameters. Recent approaches to domain adversarial training in segmentation models have used multi-level domain confusion approaches \cite{pei2018multi, Ciga_2019} which attempt to match intermediate representations of the data at different stages of the network. This may help adapt segmentation networks better.

\begin{figure}[H] 
    \centering
    \includegraphics[width=1\linewidth]{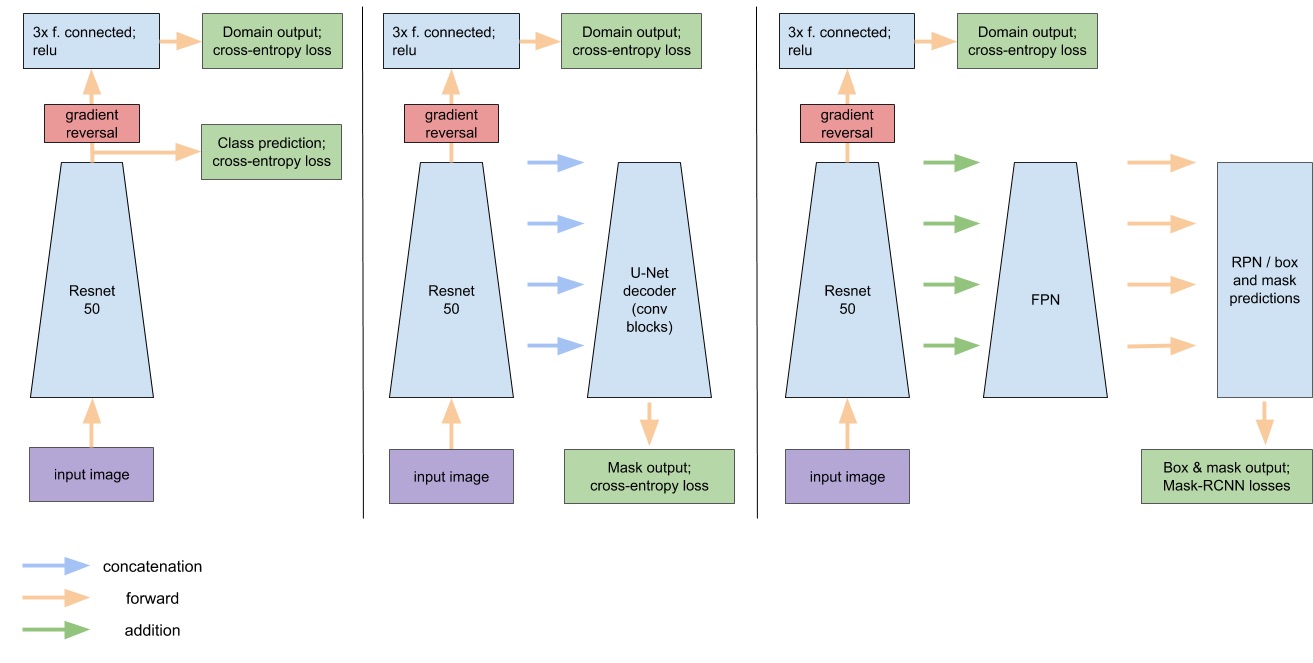}
    \caption{The three architectures we used. In all cases, we used an ImageNet pre-trained Resnet 50 as the initial encoder. The output of this is where domain adversarial training, using reverse gradients, was applied. For details on the remaining components of the architecture, we refer to \cite{DBLP:journals/corr/HeZRS15, DBLP:journals/corr/RonnebergerFB15, maskrcnn_he_2017}.}
    \label{fig:architectures}
    \vspace{-1em}
\end{figure}

\end{document}